\documentclass[runningheads]{llncs}
\usepackage{graphicx}
\usepackage{amsmath}
\usepackage{amssymb}
\usepackage{booktabs}
\usepackage{rotating}
\usepackage{makecell}
\usepackage{xspace}
\usepackage{xcolor}
\usepackage[export]{adjustbox}
\usepackage{multirow}
\usepackage{float}
\usepackage{pifont}
\usepackage{caption}
\usepackage{subcaption}
\usepackage{multirow}
\usepackage[pagebackref,breaklinks,colorlinks]{hyperref}

\newcommand{\placeholder}[1]{{\color{red}{[X]}}}

\makeatletter
\DeclareRobustCommand\onedot{\futurelet\@let@token\@onedot}
\def\@onedot{\ifx\@let@token.\else.\null\fi\xspace}

\def\etal{\emph{et al}\onedot}
\makeatother

\usepackage[capitalize]{cleveref}
\crefname{section}{Sec.}{Secs.}
\Crefname{section}{Section}{Sections}
\Crefname{table}{Table}{Tables}
\crefname{table}{Tab.}{Tabs.}
\crefname{subtable}{Fig.}{Figures}

\begin{document}
\title{GEMTrans: A General, Echocardiography-based, Multi-Level Transformer Framework for Cardiovascular Diagnosis
}
\titlerunning{Multi-Level Transformer Framework for Cardiovascular Diagnosis}
%
\author{Masoud Mokhtari\inst{1} \and
Neda Ahmadi\inst{1} \and
Teresa S. M. Tsang\inst{2} \and
Purang Abolmaesumi\inst{1}\thanks{Corresponding Authors}\and
Renjie Liao\inst{1}}
%
\authorrunning{M. Mokhtari et al.}
%
\institute{Electrical and Computer Engineering, University of British Columbia,
Vancouver, BC, Canada \\
\email{\{masoud, nedaahmadi, purang, rjliao\}@ece.ubc.ca} \and
Vancouver General Hospital, Vancouver, BC, Canada\\
\email{t.tsang@ubc.ca}}

\maketitle              

\begin{abstract}
Echocardiography (echo) is an ultrasound imaging modality that is widely used for various cardiovascular diagnosis tasks.  Due to inter-observer variability in echo-based diagnosis, which arises from the variability in echo image acquisition and the interpretation of echo images based on clinical experience, vision-based machine learning (ML) methods have gained popularity to act as secondary layers of verification. For such safety-critical applications, it is essential for any proposed ML method to present a level of explainability along with good accuracy. In addition, such methods must be able to process several echo videos obtained from various heart views and the interactions among them to properly produce predictions for a variety of cardiovascular measurements or interpretation tasks. Prior work lacks explainability or is limited in scope by focusing on a single cardiovascular task. To remedy this, we propose a \textbf{G}eneral, \textbf{E}cho-based, \textbf{M}ulti-Level \textbf{T}ransformer (GEMTrans) framework that provides explainability, while simultaneously enabling multi-video training where the inter-play among echo image patches in the same frame, all frames in the same video, and inter-video relationships are captured based on a downstream task. We show the flexibility of our framework by considering two critical tasks including ejection fraction (EF) and aortic stenosis (AS) severity detection. Our model achieves mean absolute errors of 4.15 and 4.84 for single and dual-video EF estimation and an accuracy of 96.5\% for AS detection, while providing informative task-specific attention maps and prototypical explainability.

\end{abstract}
\section{Introduction and Related Works}
\label{sec:intro}
Echocardiography (echo) is an ultrasound imaging modality that is widely used to effectively depict the dynamic cardiac anatomy from different standard views \cite{ultrasound1}. Based on the orientation and position of the obtained views with respect to the heart anatomy, different measurements and diagnostic observations can be made by combining information across several views. For instance, Apical Four Chamber (A4C) and Apical Two Chamber (A2C) views can be used to estimate ejection fraction (EF) as they depict the left ventricle (LV), while Parasternal Long Axis (PLAX) and Parasternal Short Axis (PSAX) echo can be used to detect aortic stenosis (AS) due to the visibility of the aortic valve in these views. 

Challenges in accurately making echo-based diagnosis have given rise to vision-based machine learning (ML) models for automatic predictions. A number of these works perform segmentation of cardiac chambers; Liu \etal \cite{pyramid-seg} perform LV segmentation using feature pyramids and a segmentation coherency network, while Cheng \etal \cite{contrast-seg} and Thomas \etal \cite{graph-seg} use contrastive learning and GNNs for the same purpose, respectively. Some others introduce ML frameworks for echo view classification \cite{view1,view2} or detecting important phases (e.g. end-systole (ES) and end-diastole (ED)) in a cardiac cycle \cite{phase}. Other bodies of work focus on making disease prediction from input echo. For instance, Duffy \etal \cite{echonet-lvh} perform landmark detection to predict LV hypertrophy, while Roshanitabrizi \etal \cite{rheum} predict rheumatic heart disease from Doppler echo using an ensemble of transformers and convolutional neural networks. In this paper, however, we focus on ejection fraction (EF) estimation and aortic stenosis severity (AS) detection as two example applications to showcase the generality of our framework and enable its comparison to prior works in a tractable manner. Therefore, in the following two paragraphs, we give a brief introduction to EF, AS and prior automatic detection works specific to these tasks.

EF is a ratio that indicates the volume of blood pumped by the heart and is an important indicator of heart function. The clinical procedure to estimating this ratio involves finding the ES and ED frames in echo cine series (videos) and tracing the LV on these frames. A high level of inter-observer variability of $7.6\%$ to $13.9\%$ has been reported in clinical EF estimates \cite{echonet}. Due to this, various ML models have been proposed to automatically estimate EF and act as secondary layers of verification. More specifically, Esfeh \etal \cite{bayesianef} propose a Bayesian network that produces uncertainty along with EF predictions, while Reynaud \etal \cite{transformeref} use BERT \cite{bert} to capture frame-to-frame relationships. Recently, Mokhtari \etal \cite{echognn} provide explainability in their framework by learning a graph structure among frames of an echo. 

The other cardiovascular task we consider is the detection of AS, which is a condition in which the aortic valve becomes calcified and narrowed, and is typically detected using spectral Doppler measurements~\cite{AS_medical_journal,AS_doppler}. High inter-observer variability, limited access to expert cardiac physicians, and the unavailability of spectral Doppler in many point-of-care ultrasound devices are challenges that can be addressed through the use of automatic AS detection models. For example, Huang et al. \cite{TMED1,TMED2} predict the severity of AS from single echo images, while Ginsberg et al. \cite{Tom} adopt a multitask training scheme.

\begin{figure}[t]
    \centering
    \includegraphics[width=0.85\linewidth]{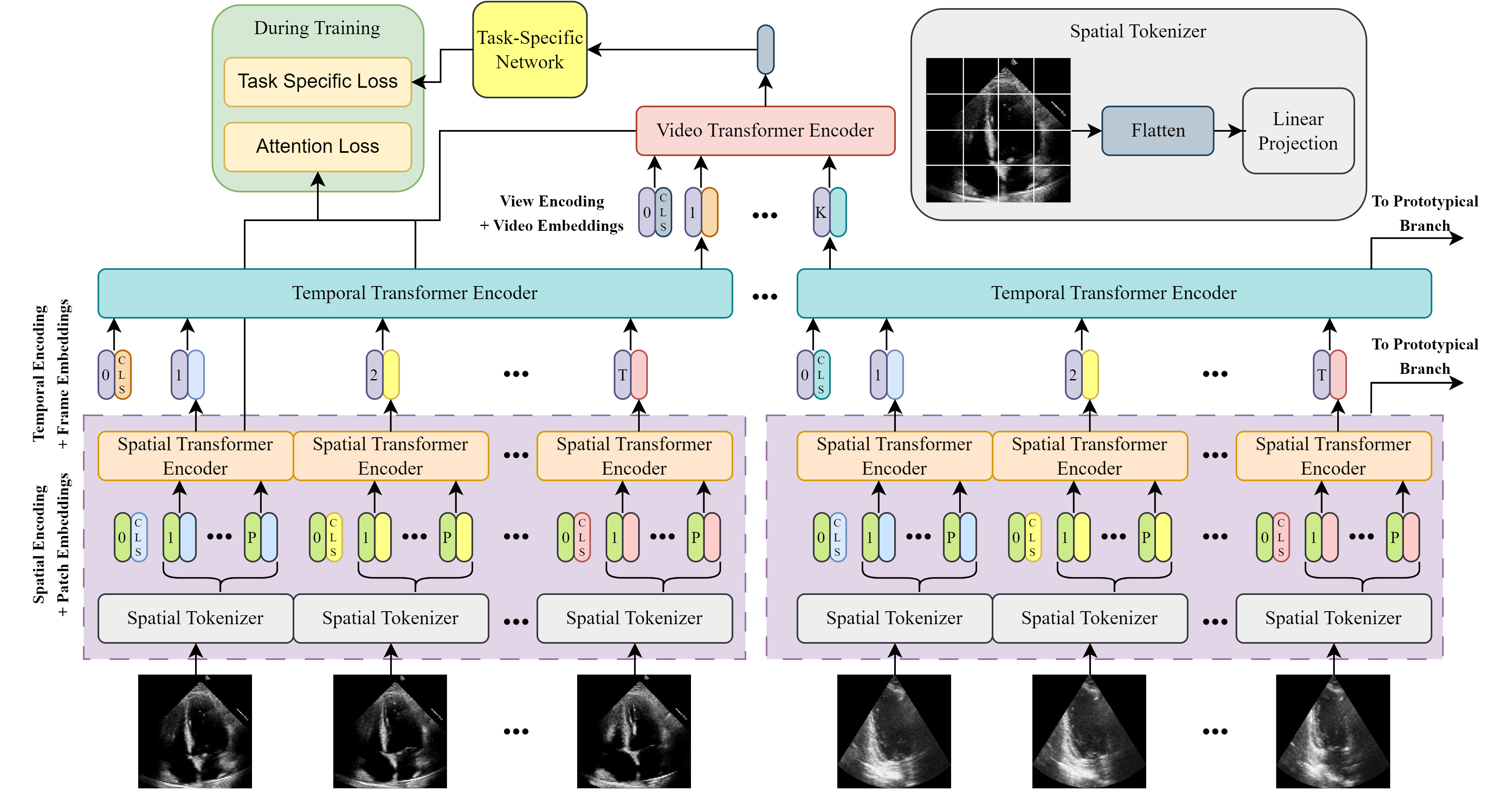}
    \caption{\textbf{GEMTrans Overview} - The multi-level transformer network processes one or multiple echo videos and is composed of three main components. \emph{Spatial Transformer Encoder (STE)} produces attention among patches in the same image frame, while \emph{Temporal Transformer Encoder (TTE)} captures the temporal dependencies among the frames of each video. Lastly, \emph{Video Transformer Encoder (VTE)} produces an embedding summarizing all available data for a patient by processing the learned embedding of each video. Different downstream tasks can then be performed using this final learned embedding. During training, both the final prediction and the attention learned by different layers of the framework can be supervised (not all connections are shown for cleaner visualization).}
    \label{fig:transformer_arch}
\end{figure}

Our framework is distinguished from prior echo-based works on multiple fronts. First, to the best of our knowledge, our model is the first to produce attention maps on the patch, frame and video levels for echo data, while allowing multiple videos to be processed simultaneously as shown in \cref{fig:transformer_arch}. Second, unlike prior works that are proposed for a single cardiovascular task, our framework is general and can be modified for a variety of echo-based metrics. More concretely, the versatility of our framework comes from its multi-level attention mechanism. For example, for EF, temporal attention between the echo frames is critical to capture the change in the volume of the LV, while for AS, spatial attention to the valve area is essential, which is evident by how the model is trained to adapt its learned attention to outperform all prior works. Lastly, we provide patch and frame-level, attention-guided prototypical explainability.

Our contributions are summarized below:
\begin{itemize}
    \item[$\bullet$] We propose GEMTrans, a general, transformer-based, multi-level ML framework for vison-based medical predictions on echo cine series (videos).
    \item[$\bullet$] We show that the task-specific attention learned by the model is effective in highlighting the important patches and frames of an input video, which allows the model to achieve a mean absolute error of 4.49 on two EF datasets, and a detection accuracy score of 96.5\% on an AS dataset.
    \item[$\bullet$] We demonstrate how prototypical learning can be easily incorporated into the framework for added multi-layer explainability.
\end{itemize}

\section{Method}
\label{sec:method}

\subsection{Problem Statement}
\label{sec:problem-statement}
The input data for both EF and AS tasks compose of one or multiple B-mode echo videos denoted by $X \in \mathbb{R}^{K\times T \times H \times W}$, where $K$ is the number of videos per sample, $T$ is the number of frames per video, and $H$ and $W$ are the height and width of each grey-scale image frame.
For \textbf{EF Estimation}, we consider both the single video (A4C) and the dual video (A2C and A4C) settings corresponding to $K=1$ and $K=2$, respectively. Our datasets consist of triplets $\{x_\text{ef}^i,y_\text{ef}^i,y_\text{seg}^i\}$, where $i \in [1,...,n]$ is the sample number. $y_\text{ef}^i \in [0,1]$ is the ground truth EF value, $y_\text{seg}^i \in \{0,1\}^{H \times W}$ is the binary LV segmentation mask, and $x^i \in \mathbb{R}^{K \times T \times H\times W}$ are the input videos defined previously.
The goal is to learn an EF estimation function $f_\text{ef}: \mathbb{R}^{K \times T \times H\times W} \mapsto \mathbb{R}$. For \textbf{AS Classification}, we consider the dual-video setting (PLAX and PSAX). Here, our dataset $D_\text{as} = \{X_\text{as},Y_\text{as}\}$ consists of pairs $\{x_\text{as}^i, y_\text{as}^i\}$, where $x^i_\text{as}$ are the input videos and $y_{\text{as}}^i \in \{0,1\}^4$ is a one-hot label indicating healthy, mild, moderate and severe AS cases. Our goal is to learn $f_\text{as}: \mathbb{R}^{2 \times T \times H\times W} \mapsto \mathbb{R}^{4}$ that produces a probability over AS severity classes.

\subsection{Multi-Level Transformer Network}
\label{sec:model-arch}
As shown in \cref{fig:transformer_arch}, we employ a three-level transformer network, where the levels are tasked with patch-wise, frame-wise and video-wise attention, respectively.

\textbf{Spatial Transformer Encoder (STE)} captures the attention among patches within a certain frame and follows ViT's \cite{vit} architecture. As shown in \cref{eq:st1,eq:st2}, the Spatial Tokenizer (ST) first divides the image into non-overlapping $p \times p$ sized patches before flattening and linearly projecting each patch:

\begin{align}
    \hat{x}_{k,t}=[\hat{x}_{k,t,1},\hat{x}_{k,t,2},...,\hat{x}_{k,t,HW/p^2}] &= f_\text{patch}(x_{k,t}, p); \label{eq:st1} \\
    x'_{k,t} = [x'_{k,t,1},x'_{k,t,2},...,x'_{k,t,HW/p^2}] &= f_{\text{lin}}(\text{flatten} (\hat{x}_{k,t})), \label{eq:st2}
\end{align}
where $p$ is the patch size, $k \in [1,...,K]$ is the video number, $t \in [1,...,T]$ is the frame number, $f_\text{patches}:\mathbb{R}^{H\times W} \mapsto \mathbb{R}^{HW/p^2 \times p \times p}$ splits the image into equally-sized patches, and $f_\text{lin}:\mathbb{R}^{p^2} \mapsto \mathbb{R}^{d}$ is a linear projection function that maps the flattened patches into $d$-dimensional embeddings. The obtained tokens from the ST are then fed into a transformer network \cite{transformer} as illustrated in \cref{eq:trans1,eq:trans2,eq:trans3,eq:trans4}:

\begin{align}
    h^0_{k,t} &= [\text{cls}_\text{spatial};x'_{k,t}] + E_\text{pos}; \label{eq:trans1} \\
    h'^l_{k,t} &=  \text{MHA}(\text{LN}(h^{l-1}_{k,t})) + h^{l-1}_{k,t}, \quad l\in[1,...,L]; \label{eq:trans2}\\
    h^l_{k,t} &= MLP(LN(h'^l_{k,t})) + h'^l_{k,t}, \quad l\in[1,...,L];  \label{eq:trans3} \\
    z_{k,t} &= LN(h^L_{k,t,0})\label{eq:trans4},
\end{align}
where $\text{cls}_\text{spatial} \in \mathbb{R}^d$ is a token similar to the $[class]$ token introduced by Devlin \etal \cite{bert}, $E_\text{pos} \in \mathbb{R}^d$ is a learnable positional embedding, MHA is a multi-head attention network \cite{transformer}, LN is the LayerNorm operation, and MLP is a multi-layer perceptron. The obtained result $z_{k,t} \in \mathbb{R}^d$ can be regarded as an embedding summarizing the $t_{th}$ frame in the $k_{th}$ video for a sample.

\textbf{Temporal Transformer Encoder (TTE)} accepts as input the learned embeddings of the STE for each video $z_{k, 1...T}$ and performs similar operations outlined in \cref{eq:trans1,eq:trans2,eq:trans3,eq:trans4} to generate a single embedding $v_k \in \mathbb{R}^d$ representing the whole video from the $k_{th}$ view. \textbf{Video Transformer Encoder (VTE)} is the same as TTE with the difference that each input token $v_k \in \mathbb{R}^d$ is a representation for a complete video from a certain view. The output of VTE is an embedding $u^i \in \mathbb{R}^d$ summarizing the data available for patient $i$. This learned embedding can be used for various downstream tasks as described in \cref{sec:optimization}.

\subsection{Attention Supervision}
\label{sec:attention_sup}

For EF, the ED/ES frame locations and their LV segmentation are available. This intermediary information can be used to supervise the learned attention of the transformer. Therefore, inspired by Stacey \etal \cite{supervise_attn}, we supervise the last-layer attention that the cls token allocates to other tokens. More specifically, for spatial attention, we penalize the model for giving attention to the region outside the LV, while for temporal dimension, we encourage the model to give more attention to the ED/ES frames. More formally, we define $\text{ATTN}^{\text{spatial}}_{\text{cls}} \in [0,1]^{HW/p^2}$ and $\text{ATTN}^{\text{temporal}}_{\text{cls}} \in [0,1]^{T}$ to be the $L_{th}$-layer, softmax-normalized spatial and temporal attention learned by the MHA module (see \cref{eq:trans2}) of STE and TTE, respectively. The attention loss is defined as

\begin{align}
    y'_{\text{seg}} &= \text{OR}(y'^{\text{ed}}_{\text{seg}},y'^{\text{es}}_{\text{seg}}); \label{eq:sup_attn_1}\\ 
L^\text{spatial}_\text{attn, s} &= 
\begin{cases}
    (\text{ATTN}^{\text{spatial}}_{\text{cls},s} - 0)^2,& \text{if } y'_{\text{seg},s} = 0 \text{ (outside LV)} \\
    0,              & \text{otherwise;}
\end{cases} 
\label{eq:sup_attn_2} \\
L^\text{temporal}_\text{attn, t} &= 
\begin{cases}
    (\text{ATTN}^{\text{temporal}}_{\text{cls},s} - 1)^2,& \text{if } t \in [\text{ED}, \text{ES}]\\
    0,              & \text{otherwise;}
\end{cases}
\label{eq:sup_attn_3} \\
L_{\text{attn}} &= \lambda_{\text{temporal}} \Sigma_{t=1}^T L^\text{temporal}_\text{attn, t} + \lambda_{\text{spatial}} \Sigma_{s=1}^{HW/p^2} L^\text{spatial}_\text{attn, s}, \label{eq:sup_attn_4}
\end{align}
where $y'^{\text{ed}}_{\text{seg}}, y'^{\text{es}}_{\text{seg}} \in \{0,1\}^{HW/p^2}$ are the coarsened versions (to match patch size) of $y_\text{seg}$ at the ED and ES locations, OR is the bit-wise logical \textit{or} function, ed/es indicate the ED/ES temporal frame indices, and $\lambda_{\text{temporal}}, \lambda_{\text{spatial}} \in [0,1]$ control the effect of spatial and temporal losses on the overall attention loss. \textit{A figure is provided in the supp. material for further clarification.}

\subsection{Prototypical Learning}
\label{sec:proto}
Prototypical learning provides explainability by presenting training examples (prototypes) as the reasoning for choosing a certain prediction. As an example, in the context of AS, patch-level prototypes can indicate the most important patches in the training frames that correspond to different AS cases. Due to the multi-layer nature of our framework and inspired by Xue \etal~\cite{Xue2022ProtoPFormerCO}, we can expand this idea and use our learned attention to filter out uninformative details prior to learning patch and frame-level prototypes. As shown in \cref{fig:transformer_arch}, the STE and TTE's attention information are used in prototypical branches to learn these prototypes. It must be noted that we are not using prototypical learning to improve performance, but rather to provide added explainability. For this reason, prototypes are obtained in a post-processing step using the pretrained transformer framework. \textit{Prototypical networks are provided in the supp. material.}

\subsection{Downstream Tasks and Optimization}
\label{sec:optimization}

The output embedding of VTE denoted by $u \in \mathbb{R}^{n \times d}$ can be used for various downstream tasks. We use \cref{eq:ef_1} to generate predictions for EF and use an L2 loss between $\hat{y}_{\text{ef}}$ and $y^i_{\text{ef}}$ for optimization denoted by $L_\text{ef}$. For AS severity classification, \cref{eq:as_1} is used to generate predictions and a cross-entropy loss is used for optimization shown as $L_\text{as}$:

\begin{align}
    \hat{y}_{\text{ef}} &= \sigma(\text{MLP}(u)) \quad \hat{y}_{\text{ef}} \in \mathbb{R}^{n\times 1}; \label{eq:ef_1} \\
    \hat{y}_{\text{as}} &= \text{softmax}(\text{MLP}(u)) \quad \hat{y}_{\text{as}} \in \mathbb{R}^{n \times 4}; \label{eq:as_1} \\
        L_{\text{overall}} &= L_\text{ef or as} + L_\text{attn}, \label{eq:loss}
\end{align}
where $\sigma$ is the Sigmoid function. Our overall loss function is shown in \cref{eq:loss}.

\section{Experiments}
\label{sec:experiments}

\subsection{Implementation}
\label{sec:implementation}
Our code-base, pre-trained model weights and the corresponding configuration files are provided at \url{https://github.com/DSL-Lab/gemtrans}. All models were trained on four 32~GB NVIDIA Tesla V100 GPUs, where the hyper-parameters are found using Weights \& Biases random sweep \cite{wandb}. Lastly, we use the ViT network from \cite{vit-imp} pre-trained on ImageNet-21K for the STE module.

\subsection{Datasets}
\label{sec:datasets}
We compare our model's performance to prior works on three datasets. In summary, for the single-video case, we use the EchoNet Dynamic dataset that consists of $10,030$ AP4 echo videos obtained at Stanford University Hospital \cite{echonet} with a training/validation/test (TVT) split of $7,465$, $1,288$ and $1,277$. For the dual-video setting, we use a private dataset of $5,143$ pairs of AP2/AP4 videos with a TVT split of $3,649$, $731$, $763$. For the AS severity detection task, we use a private dataset of PLAX/PSAX pairs with a balanced number of healthy, mild, moderate and severe AS cases and a TVT of $1,875$, $257$, and $258$. For all datasets, the frames are resized to $224 \times 224$. Our use of this private dataset is approved by the University of British Columbia Research Ethics Board, with an approval number H20-00365.

\subsection{Results}
\label{sec:results}

\begin{figure}
    \begin{subtable}[h]{0.35\textwidth}
        \centering
    \begin{tabular}{c@{\hspace{1mm}}c}
        \includegraphics[width=0.42\textwidth,trim={0 0 0 0},clip,valign=m]{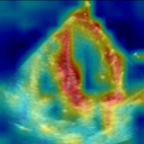} & 
        \includegraphics[width=0.42\textwidth,trim={0 0 0 0},clip,valign=m]{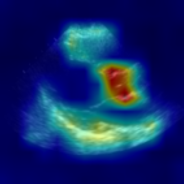} 
        \\ \addlinespace[1mm]
        \includegraphics[width=0.42\textwidth,trim={0 0 0 0},clip,valign=m]{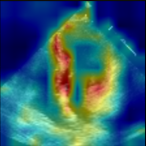} & 
        \includegraphics[width=0.42\textwidth,trim={0 0 0 0},clip,valign=m]{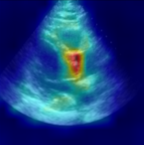}
        \\
        {\footnotesize EF} 
        & {\footnotesize AS}
        \\
    \end{tabular}
       \caption{\textbf{Learned Patch-Level Attention} - We visualize the learned attention of STE, where for EF,  the model is focusing on the walls of the LV, while for AS, the model learns to attend to the valve area, which is clinically correct.}
       \label{fig:qual_spatial}
    \end{subtable}
     \hfill
    \begin{subtable}[h]{0.6\textwidth}
        \centering
    \begin{tabular}{c@{\hspace{1mm}}c}
        \includegraphics[width=0.5\textwidth,trim={0 0 0 0},clip,valign=m]{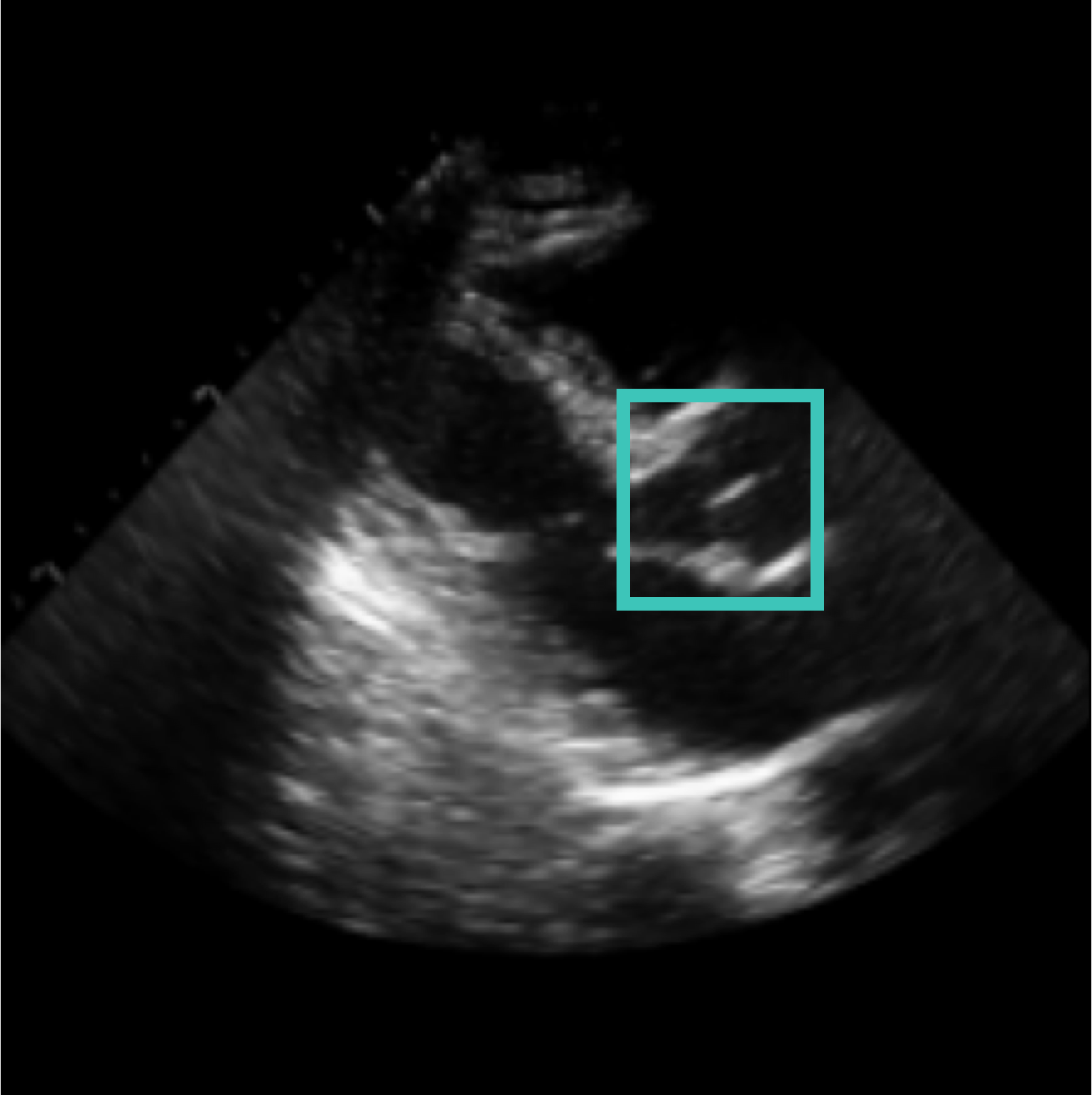} & 
        \includegraphics[width=0.5\textwidth,trim={0 0 0 0},clip,valign=m]{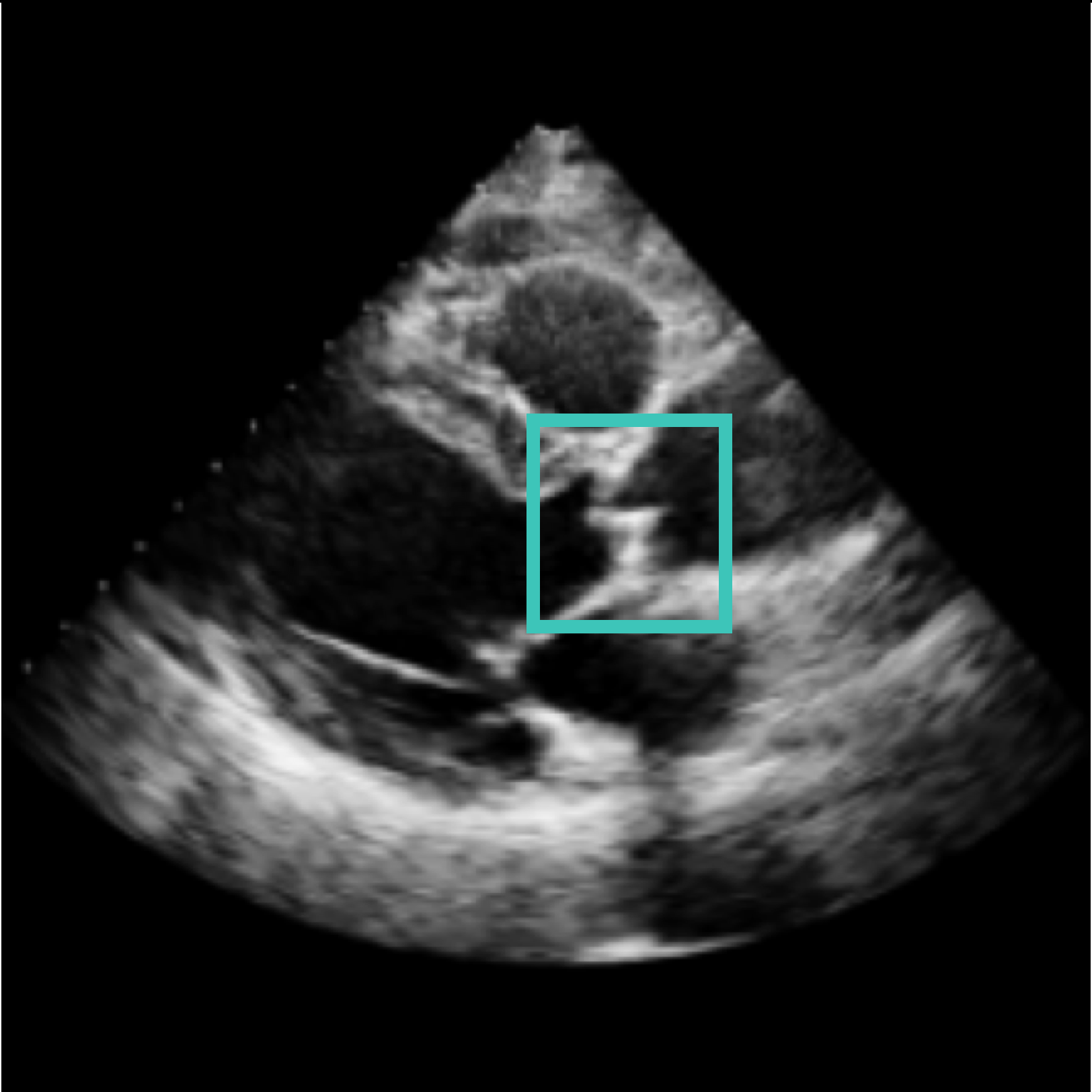}
        \\
        {\footnotesize Healthy} 
        & {\footnotesize Severe AS} 
        \\
    \end{tabular}
       \caption{\textbf{Learned Patch-Level Prototypes} - Learned prototypes use STE's attention to properly focus on the valve area for a healthy and severe AS case. We can see that in the healthy case, the aortic valve is thin and not calcified. However, in the severe case, the calcification of aortic valve is apparent (i.e. the valve appears bright in the image). \textit{Frame-level and EF prototypes presented in supp. material}. 
       }
       \label{fig:proto_vis}
    \end{subtable}
     \label{tab:temps}
     \caption{\textbf{Explainability} through learned attention and prototypes.}
\end{figure}

\textbf{Quantitative Results:} In \Cref{tab:quant_results_ef,tab:quant_results_as}, we show that our model outperforms all previous work in EF estimation and AS detection. For EF, we use the mean absolute error (MAE) between the ground truth and the predicted EF values, and $R^2$ correlation score. For AS, we use an accuracy metric for four-class AS severity prediction and binary detection of AS. The results show the flexibility of our model to be adapted for various echo-based tasks while achieving high levels of performance. \textbf{Qualitative Results:} In addition to having the superior quantitative performance, we show that our model provides explainability through its task-specific learned attention (\cref{fig:qual_spatial}) and the learned prototypes (\cref{fig:proto_vis}). \textit{More results are shown in the supp. material.} \textbf{Ablation Study:} We show the effectiveness of our design choices in \Cref{tab:ablation}, where we use EF estimation for the EchoNet Dynamic dataset as our test-bed. It is evident that our attention loss described in \cref{sec:attention_sup} is effective for EF when intermediary labels are available, while it is necessary to use a pre-trained ViT as the size of medical datasets in our experiments are not sufficiently large to build good inductive bias.

\begin{table}
\caption{\textbf{Quantitative results for EF} on the test set - LV Biplane dataset results for models not supporting multi-video training are indicated by "-". MAE is the Mean Absolute Error and $R^2$ indicates variance captured by the model.}
\label{tab:quant_results_ef}
\begin{center}
\begin{tabular}{c|cc|cc}

\multirow{2}{*}{Model} &
\multicolumn{2}{c|}{EchoNet Dynamic}&
\multicolumn{2}{c}{LV Biplane} \\
\multicolumn{1}{c|}{} &
  \multicolumn{1}{c|}{MAE {[}mm{]} $\downarrow$} &
  \multicolumn{1}{c|}{$R^2$ Score $\uparrow$} &
  \multicolumn{1}{c|}{MAE {[}mm{]} $\downarrow$} &
  \multicolumn{1}{c}{$R^2$ Score $\uparrow$} \\ \midrule\midrule
Ouyang \etal \cite{echonet}&
  \multicolumn{1}{c|}{7.35} &
  \multicolumn{1}{c|}{0.40} &
  \multicolumn{1}{c|}{-} &
  \multicolumn{1}{c}{-} 
    \\ 
Reynaud \etal \cite{transformeref}  &
  \multicolumn{1}{c|}{5.95} &
  \multicolumn{1}{c|}{0.52} &
  \multicolumn{1}{c|}{-} &
  \multicolumn{1}{c}{-}
   \\ 
Esfeh \etal \cite{bayesianef}  &
  \multicolumn{1}{c|}{4.46} &
  \multicolumn{1}{c|}{0.75} &
  \multicolumn{1}{c|}{-} &
  \multicolumn{1}{c}{-}
     \\ 
Thomas \etal \cite{graph-seg} & 
  \multicolumn{1}{c|}{4.23} &
  \multicolumn{1}{c|}{\textbf{0.79}} &
  \multicolumn{1}{c|}{-} &
  \multicolumn{1}{c}{-} \\
Mokhtari \etal \cite{echognn}  &
  \multicolumn{1}{c|}{4.45} &
  \multicolumn{1}{c|}{0.76} &
  \multicolumn{1}{c|}{5.12} &
  \multicolumn{1}{c}{0.68}
     \\ 
Ours &
  \multicolumn{1}{c|}{\textbf{4.15}} &
  \multicolumn{1}{c|}{\textbf{0.79}} &
  \multicolumn{1}{c|}{\textbf{4.84}} &
  \multicolumn{1}{c}{\textbf{0.72}}
\\
\end{tabular}
\end{center}
\end{table}

\begin{table}
\caption{\textbf{Quantitative results for AS} on the test set - \textit{Severity} is a four-class classification task, while \textit{Detection} involves the binary detection of AS.}
\label{tab:quant_results_as}
\begin{center}
\begin{tabular}{c|c|c}
\multirow{2}{*}{Model} &
\multicolumn{2}{c}{Accuracy {[}\%{]} $\uparrow$}\\
&
Severity  &
Detection \\
\midrule\midrule
Huang \etal \cite{TMED2} & 
73.7 &
94.1 \\
Bertasius \etal \cite{timesformer} & 
75.3 &
94.8 \\
Ginsberg \etal \cite{Tom} & 
74.4 &
94.2 \\
Ours & 
\textbf{76.2} &
\textbf{96.5}
\end{tabular}
\end{center}
\end{table}

\begin{table}
\caption{\textbf{Ablation study} on the validation set of EchoNet Dynamic - We see that both spatial and temporal attention supervision are effective for EF estimation, while the model does not converge without pretraining the ViT. MAE is the Mean Absolute Error and $R^2$ indicates variance captured by the model.}
\label{tab:ablation}
\begin{center}
\begin{tabular}{l|c|c}
Model &
MAE {[}mm{]} $\downarrow$ &
$R^2$ Score $\uparrow$ \\
\midrule\midrule
No Spatial Attn. Sup. & 
4.42 &
0.77 \\
No Temporal Attn. Sup. & 
4.54 &
0.76 \\
No ViT Pretraining &
5.61 &
0.45 \\
Ours & 
\textbf{4.11} &
\textbf{0.80} 
\end{tabular}
\end{center}
\end{table}

\section{Conclusions and Future Work}
\label{sec:conclusions}
In this paper, we introduced a multi-layer, transformer-based framework suitable for processing echo videos and showed superior performance to prior works on two challenging tasks while providing explainability through the use of prototypes and learned attention of the model. Future work will include training a large multi-task model with a comprehensive echo dataset that can be disseminated to the community for a variety of clinical applications.

\bibliographystyle{splncs04}
\bibliography{egbib}

\title{Supplementary Material}
%
%
\author{Masoud Mokhtari et al.}
\authorrunning{M. Mokhtari et al.}
%
\institute{Electrical and Computer Engineering, University of British Columbia,
Vancouver, BC, Canada}
\maketitle              


\begin{figure}[h!]
    \centering
    \includegraphics[width=1.0\linewidth]{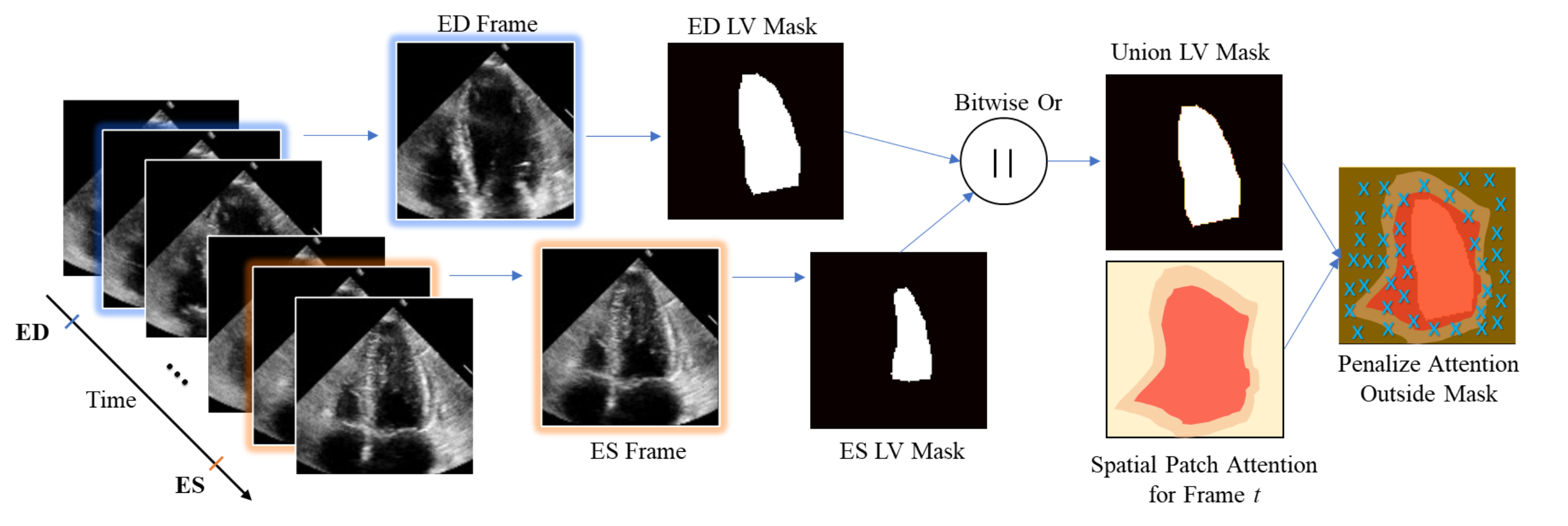}
    \caption{\textbf{Attention Supervision} - For EF, the spatial attention is penalized if attention is given outside the union of the LV mask for ED and ES. The temporal attention is also encouraged to give more attention to ED/ES locations.}
    \label{fig:attn_sup_1}
\end{figure}

\begin{figure}[h!]
    \centering
    \includegraphics[width=1.0\linewidth]{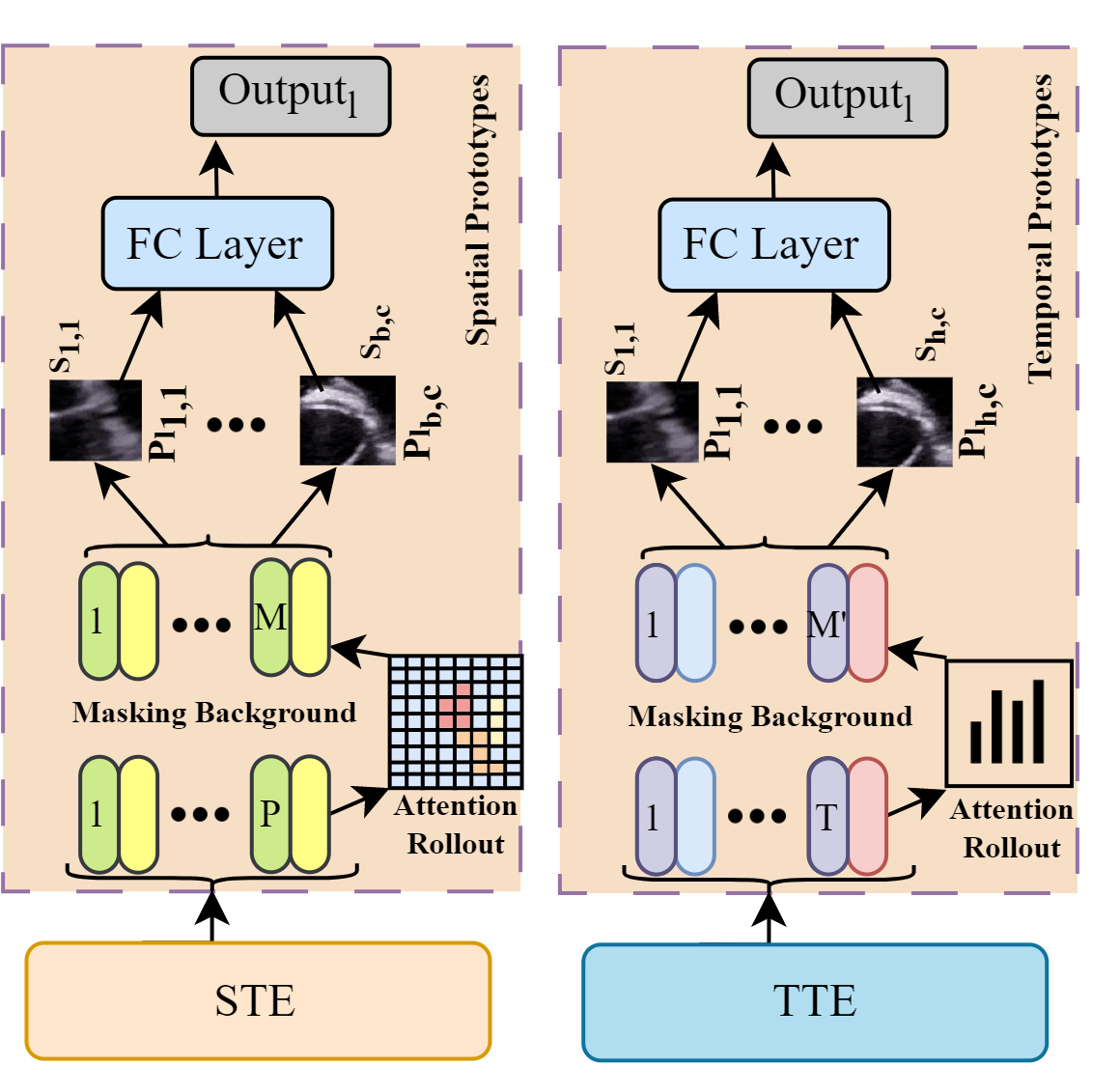}
    \caption{\textbf{Prototypical Network} -  For spatial (patch-level) prototypes, the learned local tokens $z_{k,t} \in \mathbb{R}^{HW/p^2 \times d}$ of STE are used. M patches with the highest attention are included and the rest are eliminated. Remaining patches are compared with B learnable prototypes for each class $P_l = \{p_{l_{b,c}}\}_{b=1,c=1}^{B_l,C} \in \mathbb{R}^d$ producing a similarity vector $s \in \mathbb{R}^{B\times C}$ where C is the number of classes. Fully connected layers map these similarity scores to the output. For temporal prototypes, the frame-level tokens $z'_{k,t}$ of TTE are given as input. M' frames with high temporal attention are kept and compared with H learnable prototypes and the similarity scores produce the output.}
    \label{fig:attn_sup_2}
\end{figure}

\begin{table}[h!]
    \begin{subtable}[h]{0.48\textwidth}
        \centering
    \settowidth\rotheadsize{Example 1}
    \begin{tabular}{@{\hspace{0mm}}c@{\hspace{1mm}}c@{\hspace{1mm}}c}
        \rothead{\centering Healthy} &
        \includegraphics[width=0.45\textwidth,trim={0 0 0 0},clip,valign=m]{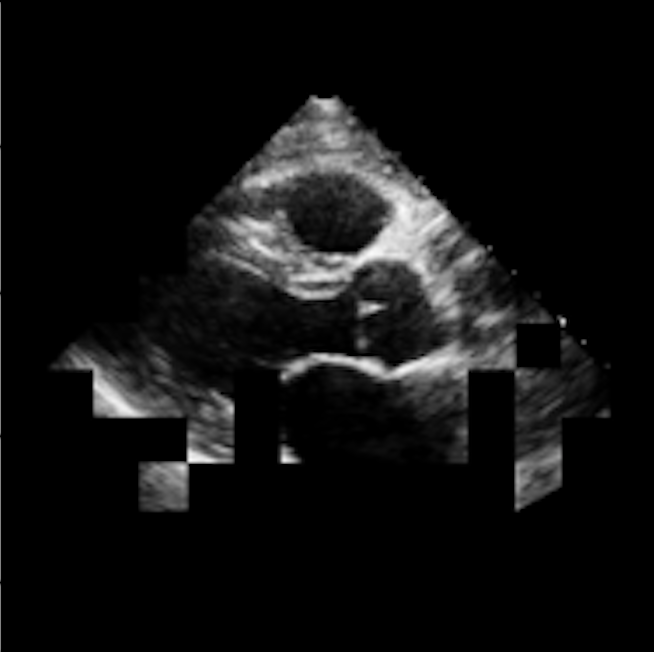} &         
        \includegraphics[width=0.45\textwidth,trim={0 0 0 0},clip,valign=m]{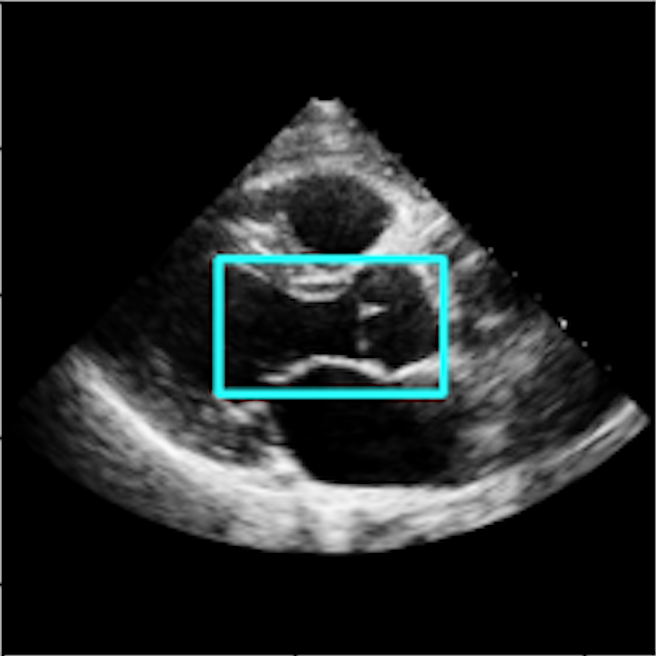} 
        \\ \addlinespace[0.5mm]
        \rothead{\centering Severe} &
        \includegraphics[width=0.45\textwidth,trim={0 0 0 0},clip,valign=m]{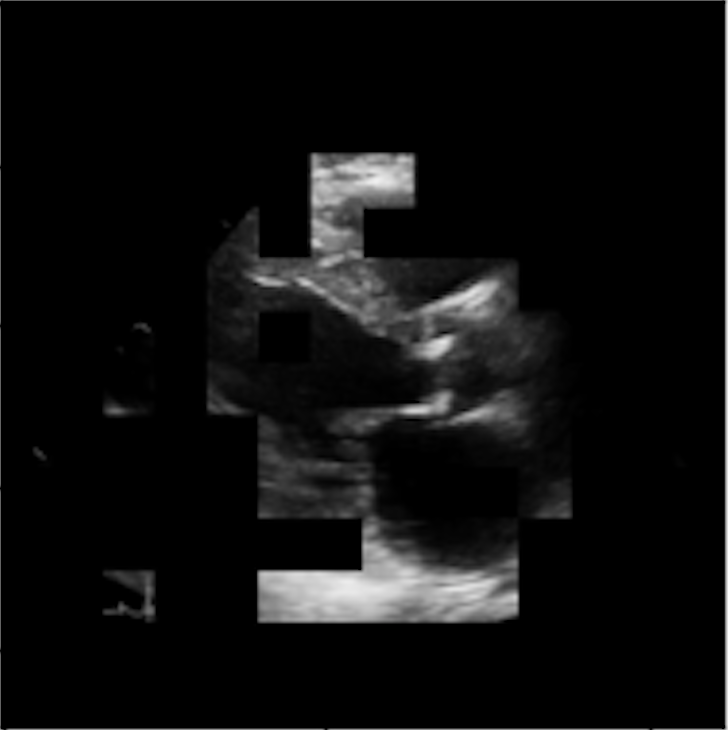} &         
        \includegraphics[width=0.45\textwidth,trim={0 0 0 0},clip,valign=m]{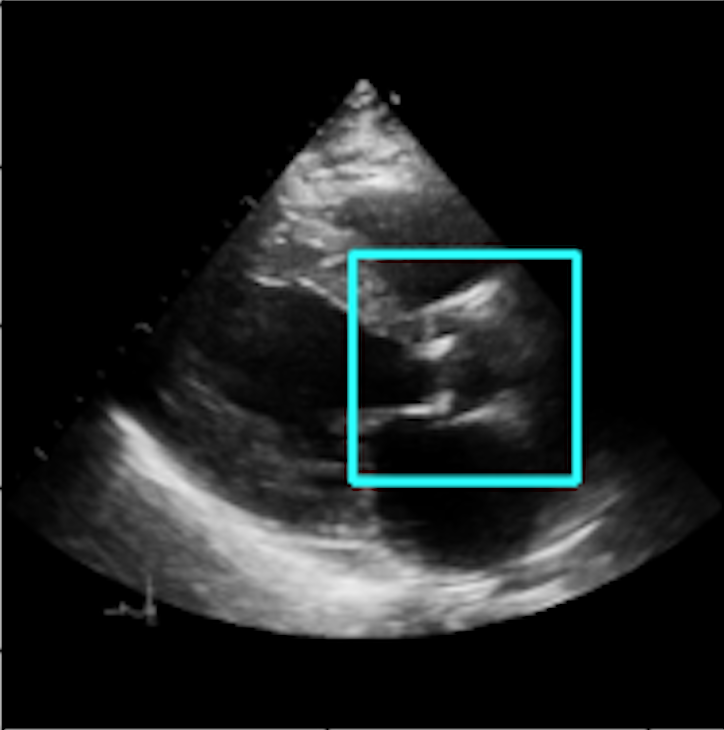} 
    \end{tabular}
    \caption*{Fig. 3: \textbf{Additional Patch-Level Prototypes for AS} - Left figures demonstrate discarded patches based on the acquired attention of STE. Patches with low attention are eliminated. The right figures display the areas that correspond to the learned prototypes. In both the healthy and severe cases, there is a notable emphasis on the aortic valve.} 
    \label{fig: qual_results_sucess_ood_2} 
    \end{subtable}
     \hfill
    \begin{subtable}[h]{0.48\textwidth}
        \centering
    \settowidth\rotheadsize{Example 1}
    \begin{tabular}{@{\hspace{0mm}}c@{\hspace{1mm}}c@{\hspace{1mm}}c}
        \rothead{\centering Healthy} &
        \includegraphics[width=0.45\textwidth,trim={0 0 0 0},clip,valign=m]{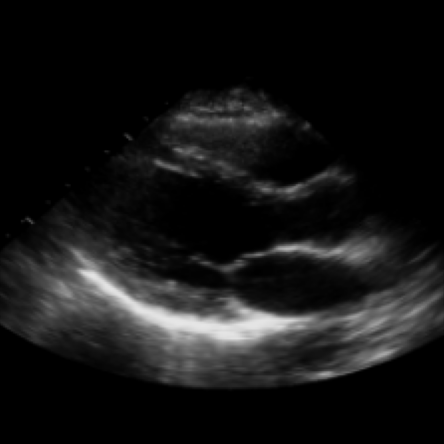} &         
        \includegraphics[width=0.45\textwidth,trim={0 0 0 0},clip,valign=m]{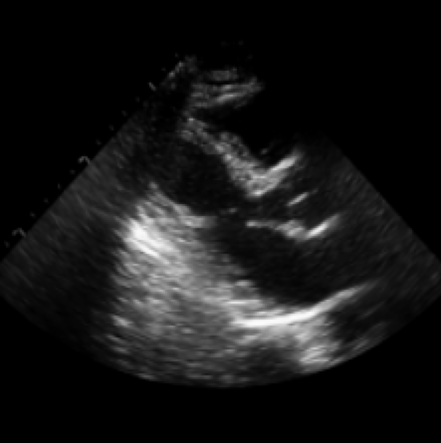} 
        \\ \addlinespace[0.5mm]
        \rothead{\centering Severe} &
        \includegraphics[width=0.45\textwidth,trim={0 0 0 0},clip,valign=m]{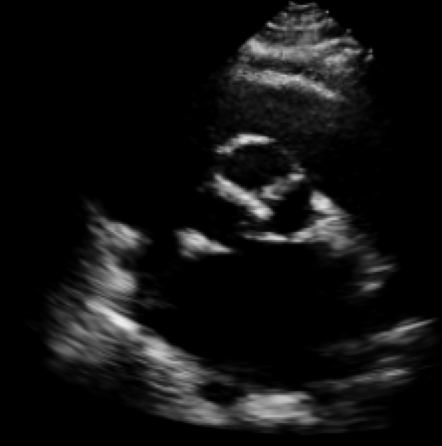} &         
        \includegraphics[width=0.45\textwidth,trim={0 0 0 0},clip,valign=m]{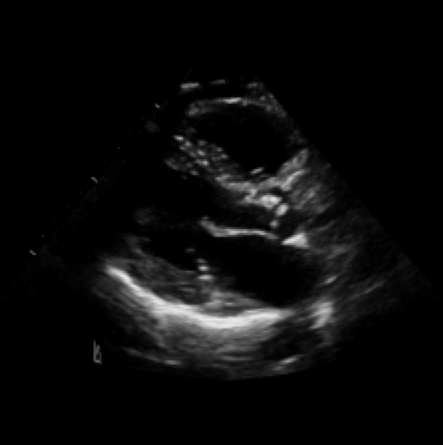} 
        \\
    \end{tabular}
    \caption*{Fig. 4: \textbf{Frame-Level Prototypes for AS} - Two instances of frame-level prototypes are visualized for healthy and severe AS. The majority of frame-level prototypes are indicative of end-systole and mid-systole stage of the heart cycle in which the restriction of valve's motion and detection of the aortic valve's calcification is easier.} 
    \label{fig: qual_results_sucess_ood_3} 
    \end{subtable}
     \label{tab:temps_1}
\end{table}

\begin{figure}[h]
    \centering
\settowidth\rotheadsize{Example 1}
\begin{tabular}{c@{\hspace{1mm}}c@{\hspace{1mm}}c}
    \includegraphics[width=0.45\textwidth,trim={0 0 0 0},clip,valign=m]{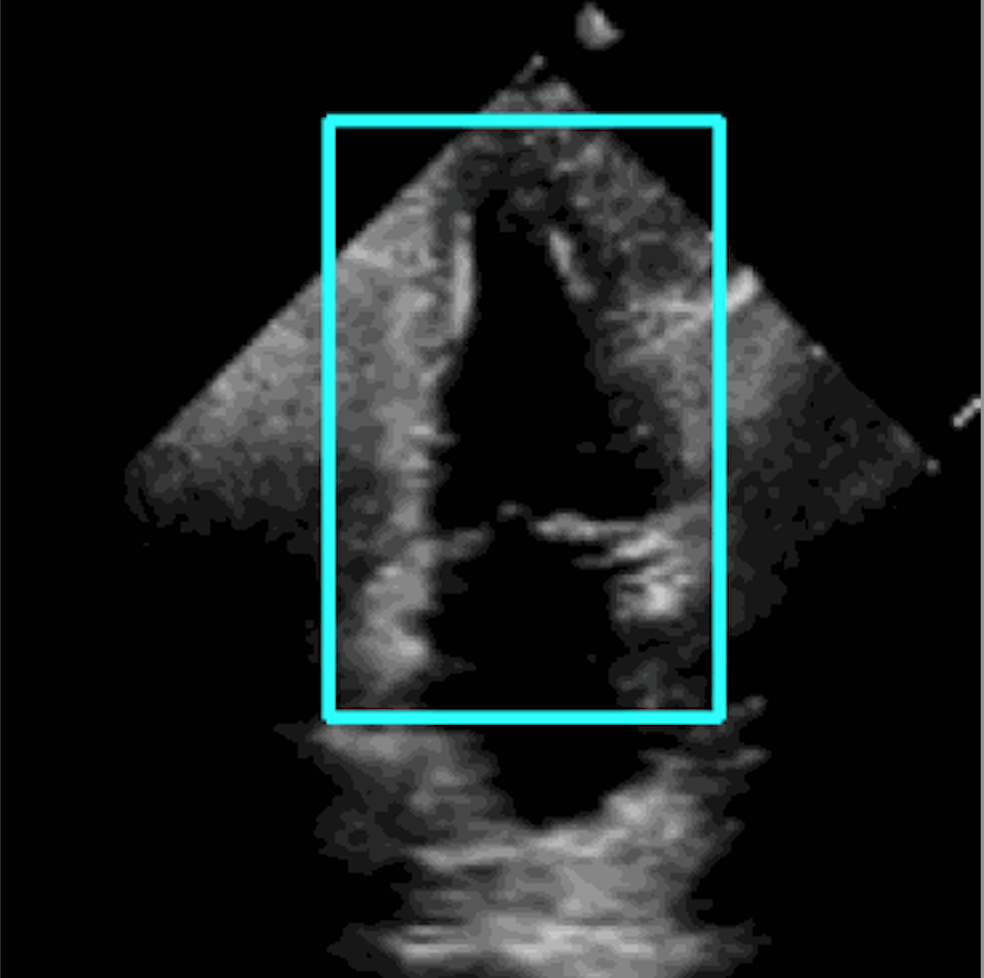} &
    \includegraphics[width=0.45\textwidth,trim={0 0 0 0},clip,valign=m]{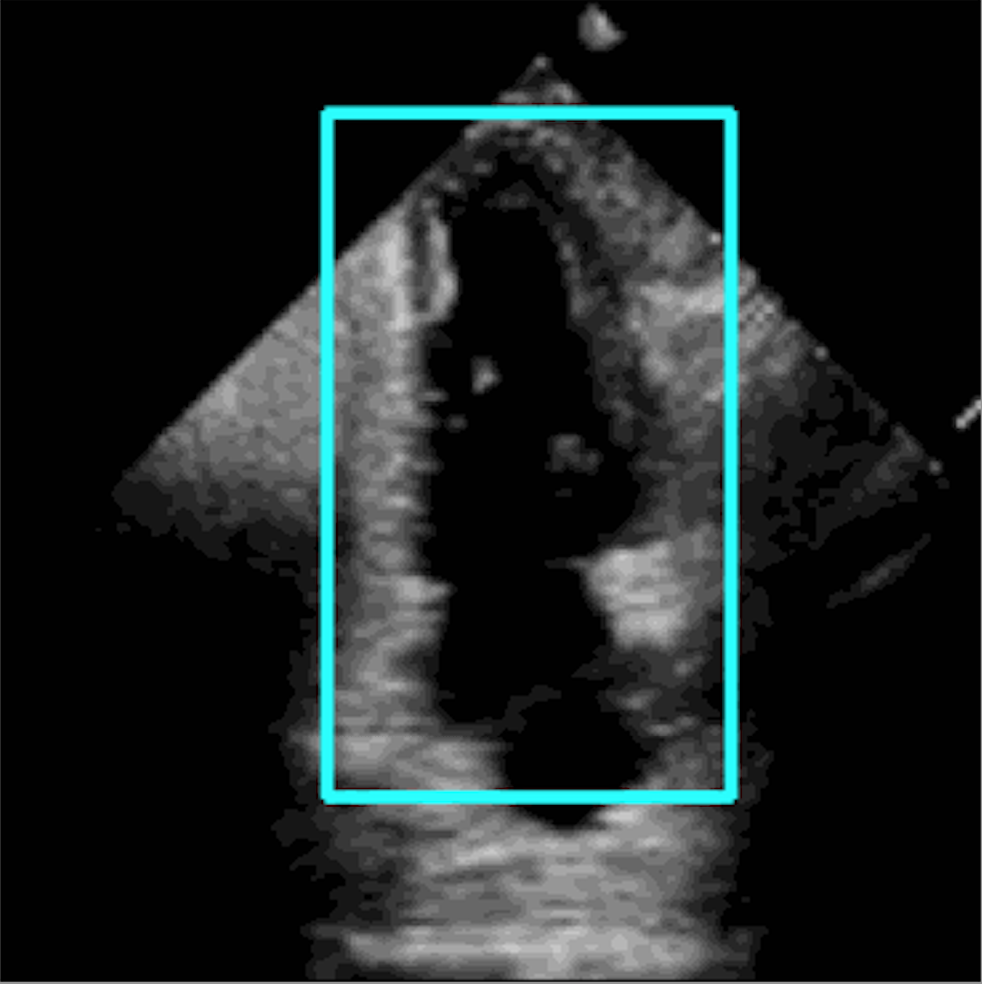} 
    \\
     {\footnotesize End-Systolic} 
        & {\footnotesize End-Dystolic }
        \\
\end{tabular}
\caption*{Fig. 5: \textbf{Patch-Level Prototypes for EF} - This figure visualizes the patch-level prototypes that represent the LV in ES and ED frames. This suggests that these frames are the most significant in contributing to the final estimation of EF, which is clinically correct since the ratio of the volume of blood in ED and ES are used to find EF.} 
\label{fig: qual_results_sucess_ood_1} 
\end{figure}

\end{document}